\documentclass[runningheads]{llncs}
\usepackage[T1]{fontenc}
\usepackage{graphicx}
\usepackage[year=2025,ID=122]{iciap}
\usepackage{graphicx}
\usepackage{caption}
\usepackage{subcaption}
\usepackage{tikz}
\usepackage{pgfplots}
\usetikzlibrary{spy,calc}
\usepackage{hyperref}
\usepackage{amsmath}
\usepackage{dblfloatfix}    
\usepackage{float}

\newcommand{\hide}[1]{}

\begin{document}
\title{Prompt-based Consistent Video Colorization}
%
%
\author{Silvia Dani\inst{1}\orcidID{0009-0002-1656-3628} \and
Tiberio Uricchio\inst{2}\orcidID{0000-0003-1025-4541} \and
Lorenzo Seidenari\inst{1}\orcidID{0000-0003-4816-0268}}
\authorrunning{Silvia Dani et al.}
%
\institute{University of Firenze, Italy \\
\email{\{name.surname\}@unifi.it}
\and
University of Pisa, Italy \\
\email{\{name.surname\}@unipi.it}\\
}
\maketitle              
\begin{abstract}
Existing video colorization methods struggle with temporal flickering or demand extensive manual input. We propose a novel approach automating high-fidelity video colorization using rich semantic guidance derived from language and segmentation. We employ a language-conditioned diffusion model to colorize grayscale frames. 
Guidance is provided via automatically generated object masks and textual prompts; our primary automatic method uses a generic prompt, achieving state-of-the-art results without specific color input. 
Temporal stability is achieved by warping color information from previous frames using optical flow (RAFT); a correction step detects and fixes inconsistencies introduced by warping. Evaluations on standard benchmarks (DAVIS30, VIDEVO20) show our method achieves state-of-the-art performance in colorization accuracy (PSNR) and visual realism (Colorfulness, CDC), demonstrating the efficacy of automated prompt-based guidance for consistent video colorization.
\end{abstract}

\section{Introduction}\label{sec:introduction}
Video colorization represents an important area of research with applications spanning historical footage restoration, media production, and artistic content creation. While significant advances have been made in this domain, existing techniques face several key challenges that limit their practical applications. Traditional colorization approaches rely heavily on user-provided color hints for grayscale frames, which can be both time-consuming to generate and often result in inconsistent colorization across frames. Furthermore, maintaining temporal consistency across colorized frames remains a significant hurdle, with independently colorized frames frequently resulting in abrupt changes and visually disruptive flickering effects.

In this paper, we explore a novel approach that leverages neural-language based methods for video colorization. Our key insight is that color information can be effectively guided through a combination of grayscale frames, automatically generated object masks, and textual prompts. Our primary automatic approach leverages a generic prompt, eliminating the need for specific color input, while we also explore the use of detailed descriptions for enhanced control.

Existing video colorization methods can be broadly categorized into reference-based approaches, which propagate colors from a reference frame throughout the video, and automatic colorization methods, which colorize each frame independently. Reference-based methods like FAVC \cite{lei2019fully} and TCVC \cite{liu2021temporally} are effective when suitable reference frames are available but struggle with scene changes and with long-term consistency. Automatic methods such as those by Zhang et al. \cite{Zhang2016ColorfulColorization} and Iizuka et al. \cite{Iizuka2016LetColor}, coming mainly from the image colorization domain, provide frame-by-frame colorization but often produce temporally inconsistent results.

Our method addresses several key limitations of existing techniques:
\begin{itemize}
    \item We eliminate the need for manual color annotation by leveraging language models for guidance, achieving state-of-the-art automatic colorization even with generic prompts.
    
	\item We achieve temporal consistency through a dual approach: using an optical flow-based method to propagate color information between frames while implementing a correction mechanism to address areas with significant color deviation.
	\item Extensive experiments on the DAVIS30 \cite{perazzi2016benchmark} and VIDEVO20 \cite{lai2018learning} datasets demonstrate that our approach achieves superior performance compared to existing colorization methods, both in terms of colorization quality and temporal consistency.
\end{itemize}

Our method is particularly beneficial for colorizing common footage where consistent and semantically accurate colors are desired without requiring extensive manual intervention. The textual prompt-based approach enables intuitive control over the colorization process, allowing for easy specification of desired color schemes through natural language.

The remainder of this paper is organized as follows: Section 2 reviews related work in video and image colorization. Section 3 details our proposed method, including the colorization pipeline, textual prompt generation, object mask creation, and temporal consistency maintenance. Section 4 presents our experimental results, including comparisons with state-of-the-art methods and ablation studies. Finally, Section 5 concludes the paper with our final remarks.
\section{Related works}\label{sec:relatedworks}
In this section, we review relevant literature in both video and image colorization, with a particular focus on approaches that inform our proposed method.
\subsection{Video Colorization}
Video colorization methods are broadly reference-based or automatic. Reference-based methods \cite{welsh2002transferring, vondrick2018tracking} propagate color from keyframes but struggle with long-term consistency and scene changes. FAVC \cite{lei2019fully} constructs KNN graphs on ground-truth color videos with temporal losses for consistency. TCVC \cite{liu2021temporally} employs bidirectional feature propagation to enhance consistency, though challenges with complex motion persist. Welsh et al. \cite{welsh2002transferring} exploit luminance features using a consistent colorized target image, identifying corresponding pixels in luminance space. Vondrick et al. \cite{vondrick2018tracking} developed self-supervised tracking via video colorization, transferring colors from reference frames. Their pointing mechanism reduces inconsistencies but occasionally produces unclear color boundaries. Liu et al. \cite{liu2021temporally}'s TCVC combines a colorization network with temporal coherence mechanisms, though still faces challenges with significant motion and occlusion. Automatic methods \cite{zhang2019deep} typically extend image colorization techniques but often lack robust temporal handling.

\subsection{Image Colorization}
While our focus is on video colorization, our approach builds upon advancements in image colorization. Early learning-based approaches treat colorization as a prediction problem. Iizuka et al. \cite{Iizuka2016LetColor} merged local patch information with global priors for complex scene colorization. Zhang et al. \cite{Zhang2016ColorfulColorization} formulated colorization as classification, predicting color distributions to capture the multimodal nature of the problem. Su et al. \cite{Su2020InstanceAware} introduced an instance-aware model (InsColor) combining object detection and colorization, demonstrating that segmentation information improves quality, particularly for distinct objects. This insight informs our use of SAM \cite{kirillov2023segment} for creating object masks.
Language-based image colorization has emerged as a powerful paradigm for user control. Manjunatha et al. \cite{Manjunatha2018ColorLanguage} pioneered using textual descriptions to guide colorization. Recent works include L-CoDe by Weng et al. \cite{Weng2022LCoDe} using color-object decoupled conditions, Chang et al. \cite{Chang2022LCoDer} transformer-based L-CoDer, and Chang et al. \cite{chang2023lcad} diffusion priors for language-based colorization.
Transformer-based approaches include Kumar et al. \cite{Kumar2021ColorizationTransformer} Colorization Transformer leveraging attention for long-range dependencies, Weng et al. \cite{weng2022ct2} CT$^2$ using color tokens, and Ji et al. \cite{Ji2022ColorFormer} ColorFormer employing hybrid-attention and color memory.

\subsection{Temporal Consistency in Video Processing}
Ensuring temporal consistency is a critical challenge in video colorization. Optical flow has been widely used to enforce consistency by warping frames or features. The RAFT architecture by Teed and Deng \cite{teed2020raft} provides accurate flow estimation for various video processing tasks. Methods like TCVC \cite{liu2021temporally} propagate deep features between frames, proving more robust than direct color propagation, especially with occlusions and large motions. Lei and Chen \cite{lei2019fully} introduced self-regularization terms that penalize discrepancies between adjacent frames, reducing flickering without explicit correspondence matching.

Our work builds upon these advances by combining language-guided colorization with instance-aware segmentation and effective temporal consistency mechanisms. We automatically generate textual descriptions at key points, segment objects using SAM, and maintain temporal consistency through a combination of optical flow and targeted correction of problematic regions.

\begin{figure}
    \centering
    \includegraphics[width =.8 \linewidth]{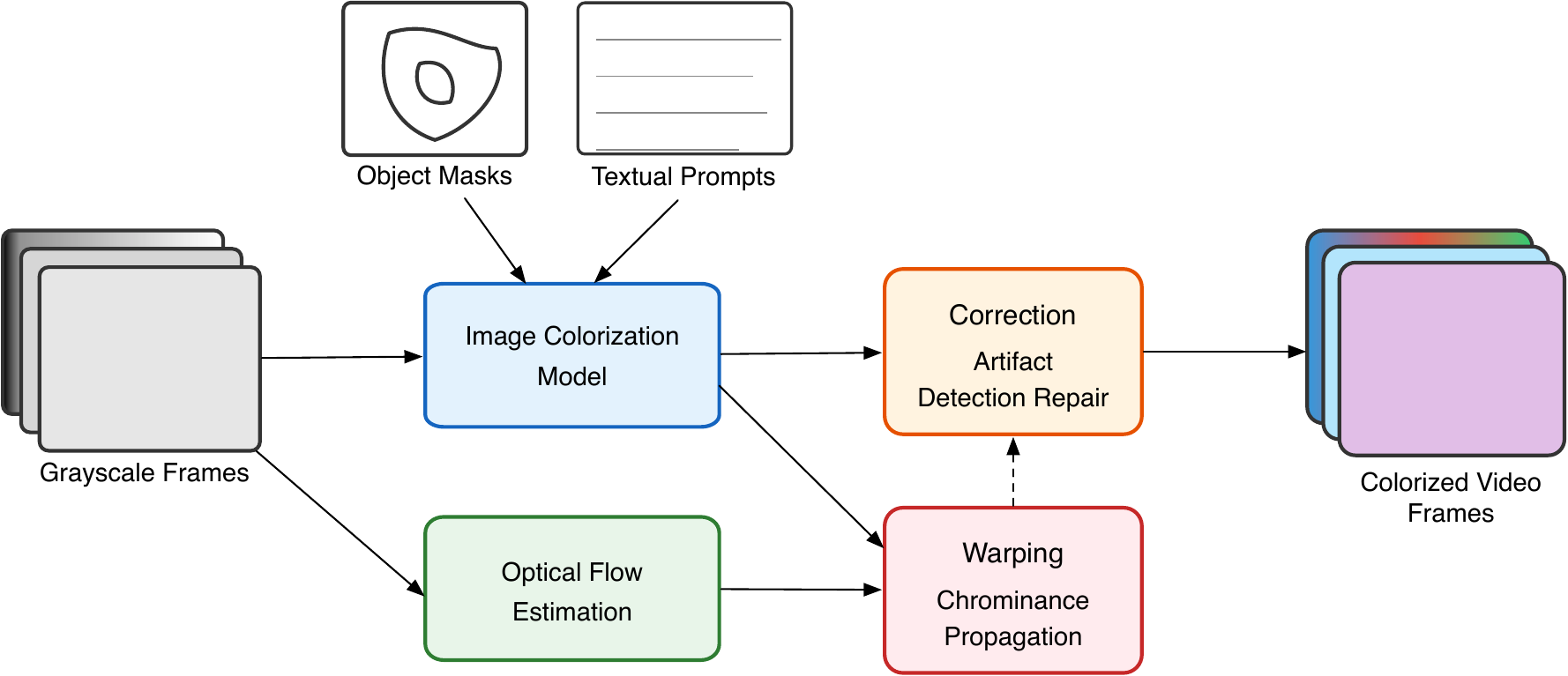}
    \caption{Overview of the proposed framework. Our method takes grayscale video frames as input, along with automatically generated object masks from SAM and a textual prompt (generic for automatic mode, detailed for guided mode).}
    \label{fig:overview-model}
\end{figure}
\vspace{-10pt}
\section{Method}\label{sec:method}
Our proposed method achieves temporally consistent video colorization guided by textual prompts and object segmentation masks. The core idea is to colorize grayscale video frames using a language-conditioned image colorization model, while enforcing temporal coherence through optical flow-based color propagation and a targeted correction mechanism. The process operates primarily in the CIE LAB color space, leveraging the separation of luminance (L) and chrominance (A, B channels). An overview of the pipeline for processing frame $t+1$ given frame $t$ is illustrated in Figure \ref{fig:overview-model}.

\subsection{Overview}
Given an input grayscale video sequence $\{L_t\}$, where $L_t$ is the luminance channel of the frame at time $t$, our goal is to generate a corresponding colorized sequence $\{I_{final}(t)\}$. For each frame $L_t$, we automatically generate a set of object masks $M_t$. We then utilize a textual prompt $P_t$ to guide the colorization. For our main automatic approach evaluated against the state-of-the-art, we use a generic prompt (e.g., "a colorful image") to ensure no external color information biases the comparison. We also explore the use of detailed, content-specific prompts to demonstrate the framework's controllability.

These inputs ($L_t$, $M_t$, $P_t$) are fed into a language-guided image colorization network (L-CAD \cite{chang2023lcad}) to produce an initial colorized frame $I_{L-CAD}(t)$. To ensure temporal consistency, we compute the optical flow $F_{t \to t+1}$ between $L_t$ and $L_{t+1}$ using RAFT \cite{teed2020raft}. We then warp the chrominance channels ($A_t, B_t$) of the previous final colorized frame $I_{final}(t)$ using $F_{t \to t+1}$ to obtain warped chrominance channels ($A'_{t+1}, B'_{t+1}$). 
These are combined with the input luminance $L_{t+1}$ to form a preliminary warped frame $I_{warp}(t+1)$. A correction step identifies regions where warping failed or introduced significant artifacts by comparing $I_{warp}(t+1)$ with $I_{final}(t)$.
These regions are filled using the colors from the L-CAD output for the current frame, $I_{L-CAD}(t+1)$, resulting in the final colorized frame $I_{final}(t+1)$.

\subsection{Language-Guided Frame Colorization}
We employ the L-CAD model \cite{chang2023lcad} as our core per-frame colorization engine. L-CAD is a diffusion-based model capable of generating diverse and controllable colorization based on textual descriptions of varying detail levels. For each grayscale input frame $L_t$, L-CAD takes the following as input: i) the grayscale frame $L_t$; ii) a textual prompt $P_t$; iii) a set of automatically generated object segmentation masks $M_t$.
The model leverages these inputs to predict the corresponding chrominance channels ($A_{L-CAD}(t), B_{L-CAD}(t)$). When provided with a \emph{generic prompt} like "a colorful image", L-CAD infers plausible colors primarily based on the grayscale structure ($L_t$), the spatial guidance from masks ($M_t$), and its internal knowledge about typical object colors learned during pre-training. When provided with detailed prompts containing color information, L-CAD uses this explicit guidance to achieve more specific colorization. The result is the colorized frame $I_{L-CAD}(t)$ in the LAB space.

Effective guidance for L-CAD involves both spatial and semantic information. 

For spatial information, we utilize the Segment Anything Model (SAM) \cite{kirillov2023segment} to automatically generate instance segmentation masks for each grayscale frame $L_t$. SAM produces highly accurate masks for various objects without requiring class labels. These masks $M_t$ are provided as input to L-CAD to help delineate object boundaries and assign colors accurately within detected regions. While SAM can generate many masks, we currently utilize all generated masks in our experiments, which was found to be optimal.

\begin{figure}
    \centering
    \begin{subfigure}[b]{0.45\textwidth}
        \centering
        \includegraphics[width=.7\textwidth]{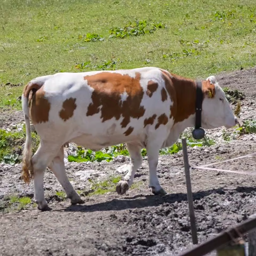}
        \caption{}
        \label{fig:cow}
    \end{subfigure}
    \hfill
    \begin{subfigure}[b]{0.45\textwidth}
        \centering
        \includegraphics[width=.7\textwidth]{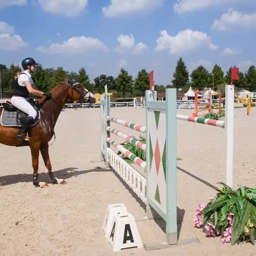}
        \caption{}
        \label{fig:horsejump-high}
    \end{subfigure}
    \caption{Textual prompt generated using ground truth color information to simulate ideal guidance from image \ref{fig:cow}: \small \textit{Cow brown and white in the center.  Grass green in the background.  Soil dark brown in the foreground.  Fence dark brown at the bottom.}\\
    Textual prompt from image \ref{fig:horsejump-high}: \small \textit{Horse is brown and is on the left.  Rider is wearing white and is on the horse.  Fence is white and is in the center.  Ground is beige and covers the bottom.  Trees are green and are in the background.  Sky is blue and is at the top.  Clouds are white and are in the sky.  Plants are green and are in the foreground.}}
\end{figure}

Regarding semantic information, we use the fixed, generic prompt $P_t = \text{"a colorful image"}$ for all frames. This ensures a fair comparison by providing no specific semantic or color guidance beyond the request for colorization itself. 
To demonstrate the potential for user control and evaluate the impact of dynamic guidance, we also utilize detailed prompts. These prompts describe scene content and specific colors (examples in Figure \ref{fig:cow}, \ref{fig:horsejump-high}). Since generating such colorful prompts automatically from grayscale input is impossible unless you have already a model with color priors, these were generated using the GPT-3.5-turbo API applied to the original ground truth color frames, simulating an oracle or ideal user providing accurate descriptions. For the dynamic prompt ablation study, a new detailed prompt is generated under two conditions: i) at the beginning of a new scene (automatically detected on the grayscale video), and ii) at regular intervals of $\Delta t = 1$ second. Between generation points, the most recent detailed prompt $P_t$ is used for subsequent frames ($P_{t+1} = P_t$).

\subsection{Temporal Consistency} 
Maintaining color stability across frames is crucial for avoiding flickering artifacts. We achieve this through a combination of optical flow-based propagation and targeted correction. We use the RAFT model \cite{teed2020raft} to estimate the dense optical flow field $F_{t \to t+1}$ between consecutive grayscale frames $L_t$ and $L_{t+1}$. The chrominance channels ($A_t, B_t$) from the final colorized frame of the previous step, $I_{final}(t)$, are warped using the backward optical flow $F_{t \to t+1}$ to predict the chrominance for the current frame $t+1$.
\begin{equation}
    (A'_{t+1}(x,y), B'_{t+1}(x,y)) = (A_t(p), B_t(p)) 
    \label{eq:warp}
\end{equation}
where $p = (x+F_{t \to t+1, x}(x,y), y+F_{t \to t+1, y}(x,y))$. Here, $(x,y)$ are coordinates in frame $t+1$, and $p$ represents the corresponding coordinates in frame $t$ according to the flow field. The warped chrominance $(A'_{t+1}, B'_{t+1})$ is combined with the input luminance $L_{t+1}$ to form the warped frame $I_{warp}(t+1)$. This step effectively transfers stable colors from the previous frame based on motion.

\begin{figure}[H]
    \centering
    \begin{subfigure}[b]{0.45\textwidth}
        \centering
        \includegraphics[width=.7\textwidth]{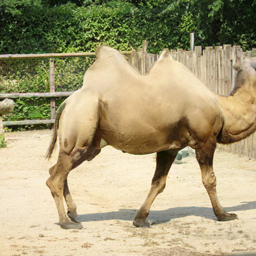}
        \caption{}
        \label{fig:first-frame}
    \end{subfigure}
    \hfill
    \begin{subfigure}[b]{0.45\textwidth}
        \centering
        \includegraphics[width=.7\textwidth]{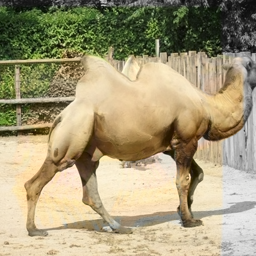}
        \caption{}
        \label{fig:color-shift}
    \end{subfigure}
    \caption{Example of color shifting from initial frame \ref{fig:first-frame} to the 11th frame \ref{fig:color-shift}.}
    \label{fig:color-shifting-example}
\end{figure}

An example of a disocclusion artifact introduced by warping is shown in Figure \ref{fig:color-shifting-example}.

If frame $t+1$ is detected as the start of a new scene, the warping step (Eq. \ref{eq:warp}) is skipped. In this case, the output from the L-CAD model for the current frame is used directly as the preliminary result: $I_{prelim}(t+1) = I_{L-CAD}(t+1)$. This prevents propagating colors across unrelated scenes.

\subsubsection{Correction of Warping Artifacts} 
Optical flow estimation and warping are not always perfect, leading to artifacts, color shifting, or uncolored (grayscale) regions, especially due to occlusions or complex motion. 
We identify these problematic areas by comparing the warped frame $I_{warp}(t+1)$ with the previous final frame $I_{final}(t)$. 
A binary mask $M_{corr}(t+1)$ identifies pixels where the distortion is significant, typically measured by a low pixel-wise Peak Signal-to-Noise Ratio (PSNR) below a threshold $\tau$:

\begin{equation}
    M_{corr}(t+1)(x,y) =
\begin{cases}
1 & \text{if } 20 \cdot \log_{10}\left(\frac{\max \left [I_{final}(t)\right]}{\|I_{warp}^{x,y}(t+1) - I_{final}^{x,y}(t)\|}\right) < \tau \\
0 & \text{otherwise}
\end{cases}
    \label{eq:corr_mask}
\end{equation}
Pixels marked by $M_{corr}(t+1)=1$ indicate regions requiring correction. 

The final colorized frame $I_{final}(t+1)$ is composed by taking the warped colors $I_{warp}(t+1)$ in regions deemed reliable ($M_{corr}=0$) and filling the corrupted regions ($M_{corr}=1$) with the colors generated by the L-CAD model for the current frame, $I_{L-CAD}(t+1)$:
\begin{equation}
    I_{final}(t+1) = (1 - M_{corr}(t+1)) \odot I_{warp}(t+1) + M_{corr}(t+1) \odot I_{L-CAD}(t+1)
    \label{eq:final_comp}
\end{equation}
where $\odot$ denotes element-wise multiplication. This selective correction ensures that we rely on propagated colors for temporal stability where possible, but fall back to the current frame's L-CAD prediction in problematic areas, maintaining both consistency and accuracy.

\subsection{Implementation Details}
Our framework integrates several pre-existing models. We use the L-CAD model \cite{chang2023lcad} with its publicly available weights as the core image colorization engine. Object masks are obtained using the default Segment Anything Model (SAM) \cite{kirillov2023segment}. 
For the main automatic method, the generic prompt ``a colorful image" is used. For the guided method, detailed prompts were generated using the GPT-3.5-turbo API from ground truth color frames, with new prompts generated at scene changes and every $\Delta t = 1$ second.
For optical flow, we employ the RAFT-Large model \cite{teed2020raft} with the weights provided by the PyTorch Vision library \cite{raft_github}. The threshold $\tau$ for detecting corrupted warping areas via PSNR (Eq. \ref{eq:corr_mask}) is set to 25 dB. All experiments were conducted using PyTorch on NVIDIA RTX 4090 GPUs.

\section{Experiments}\label{sec:experiments}
In this section, we evaluate the performance of our proposed prompt-based video colorization method. We first present quantitative and qualitative comparisons against state-of-the-art methods and then provide ablation studies to validate the contributions of key components of our framework. 

\subsection{Datasets}
We follow the experimental setup of \cite{liu2021temporally} and \cite{lei2019fully} using the test sets of DAVIS30 \cite{perazzi2016benchmark} and Videvo20 \cite{lai2018learning} datasets. Originally created for video object segmentation, the DAVIS dataset features diverse scenes with significant object motion and appearance changes. We use the standard test split containing 30 videos. Videvo20 was introduced for evaluating temporal consistency in video processing tasks. We use the test split defined in \cite{lai2018learning}, which consists of 20 videos, often characterized by smoother camera motion compared to DAVIS. As our method is training-free, we only utilize the test splits of these datasets for evaluation. 

Following existing approaches, we compare our results against the ground truth color videos using Peak Signal-to-Noise Ratio (PSNR), Colorfulness \cite{hasler2003measuring} and Color Distribution Consistency (CDC) \cite{lei2019fully}. Colorfulness is a no-reference metric quantifying the perceived richness and intensity of colors in an image. We also report the ratio of the colorfulness score of our results to that of the ground truth ($\frac{\text{Colorfulness}}{\text{True Colorfulness}}$). A ratio closer to 1 indicates a similar level of color vibrancy as the original video. CDC is also a no-reference metric designed to evaluate temporal consistency by measuring the similarity of color distributions between adjacent frames. We report the ratio relative to the ground truth ($\frac{\text{CDC}}{\text{True CDC}}$). A ratio closer to 1 signifies better temporal stability, closer to the consistency level of the original video.

\subsection{Comparison with the state of the art}
\label{sec:sota_comparison}
We compare our proposed method against several representative image and video colorization techniques. Table \ref{tab:confronto_metodi} presents the quantitative comparison on the DAVIS30 and VIDEVO20 test sets. 
We evaluate two versions of our approach: first, our main automatic approach using only the generic prompt ``a colorful image" (\emph{proposed method}). This represents a fair comparison against other automatic methods that do not use external color guidance. Second, a version that uses detailed, colorful prompts generated from ground truth (simulating ideal user guidance) to showcase the framework's potential when specific instructions are provided.

\begin{table}
\centering
\resizebox{.9\textwidth}{!}{%
    \begin{tabular}{lccc|ccc}
    \hline
     & \multicolumn{3}{c}{\textbf{DAVIS30}} & \multicolumn{3}{c}{\textbf{VIDEVO20}} \\ \hline
    {\textbf{Method}} 
    & $\frac{\text{CDC}}{\text{True CDC}}^*$ & PSNR$\uparrow$ &  $\frac{\text{Colorfulness}}{\text{True Colorfulness}}^*$   & $\frac{\text{CDC}}{\text{True CDC}}^*$  & PSNR$\uparrow$ &  $\frac{\text{Colorfulness}}{\text{True Colorfulness}}^*$ \\ \hline
    Iizuka et al. \cite{Iizuka2016LetColor} & 1.83 & 23.85 & 0.62 & 2.96 & 23.69 & 0.77 \\
    CIC \cite{Zhang2016ColorfulColorization} & 1.97 & 23.19 & 0.94 & 2.21 & 22.51 & 1.08 \\
    IDC \cite{zhang2017realtime} & 1.60 & 25.42 & 0.67 & 1.58 & 25.35 & 0.70 \\
    InsColor \cite{Su2020InstanceAware} & 3.05 & 24.51 & 0.61 & 4.92 & 24.80 & 0.75 \\ \hline
    FAVC \cite{lei2019fully} & 1.35 & 24.38 & 0.57 & 1.15 & 24.81 & 0.60 \\
    TCVC+ (b:IDC) \cite{liu2021temporally} & \textbf{1.19} & 25.50 & 0.67 & \textbf{1.01} & 25.43 & 0.70 \\ \hline
    \hline
    Proposed + prompt  & 1.25 & \textbf{29.63} & \textbf{0.94} & 1.08 & \textbf{30.26} & \textbf{1.05} \\ 
    Proposed method  & 1.36 & \textbf{28.63} & \textbf{0.93} & 1.18 & \textbf{29.54} & \textbf{1.02} \\ \hline
    \end{tabular}
}
\caption{Comparison of different methods on DAVIS30 and VIDEVO20. * Best value is closer to 1.}
\vspace{-2em}
\label{tab:confronto_metodi}
\end{table}

Our method (``Proposed method") achieves significantly higher PSNR scores than all competing methods on both datasets, indicating superior reconstruction fidelity to the ground truth colors, even without specific guidance. When provided with detailed guidance (``Proposed + prompt"), the PSNR improves further, highlighting the benefit of semantic control.

In terms of Colorfulness, both versions of our method produce results much closer to the ground truth vibrancy (ratio near 1.0) compared to most prior works (which often yield ratios < 0.7), except for CIC which can oversaturate. Regarding temporal consistency (CDC), our main automatic approach (`Proposed method`) achieves scores (1.36 on DAVIS, 1.18 on VIDEVO) reasonably close to the state-of-the-art TCVC+ (1.19 on DAVIS, 1.01 on VIDEVO), demonstrating that our flow-based propagation and correction strategy effectively maintains temporal smoothness even with generic guidance. Providing detailed prompts (`Proposed + prompt`) further improves CDC (1.25 on DAVIS, 1.08 on VIDEVO), bringing it even closer to methods specifically optimized for consistency via bidirectional feature propagation. 

\section{Ablation Studies}

\subsubsection{Impact of Dynamic Prompts}
\begin{table}[b]
\centering
\begin{tabular}{lccc}\hline

     & \multicolumn{3}{c}{\textbf{DAVIS30 pairs}} \\\hline

                               \textbf{Method} &  PSNR       $\uparrow$                   & $\frac{\text{Colorfulness}}{\text{True Colorfulness}}^*$   & $\frac{\text{CDC}}{\text{True CDC}}^*$                   \\\hline
Without Changing Prompt& 28.23& \textbf{0.98} & 1.22\\
Changing Prompt& \textbf{29.42}& 0.97& \textbf{1.15}\\\hline
\end{tabular}%
\caption{Comparison between the proposed method and its version with one fixed prompt. $^*$ Best value is closer to 1.}
\label{tab:prompt_ablation}
\end{table}

To assess the benefit of updating guidance over time when detailed information is available, we compare our ``Proposed + prompt" method (using detailed prompts generated from ground truth, updated at scene changes and every second) against a variant that uses only the detailed prompt generated for the first frame throughout the entire video. Table \ref{tab:prompt_ablation} shows the results on the DAVIS30 pairs dataset (a synthetic dataset created by concatenating random pairs from DAVIS30 to ensure each test sample contains at least one scene change).

\begin{figure}[htbp]
\centering
    \begin{subfigure}[t]{0.3\textwidth}  
        \centering
        \includegraphics[width=.7\linewidth]{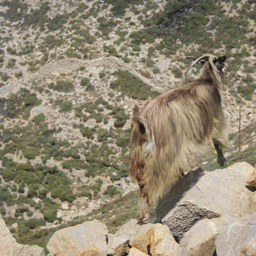}
        \caption{Changing prompt}
    \end{subfigure}
    \hfill
    \begin{subfigure}[t]{0.3\textwidth}  
        \centering  
        \includegraphics[width=.7\linewidth]{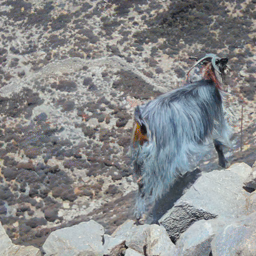}
        \caption{w/o changing prompt - content diverged}
        \label{fig:fixed_gray}
    \end{subfigure}
    \hfill
    \begin{subfigure}[t]{0.3\textwidth}  
        \centering
        \includegraphics[width=.7\linewidth]{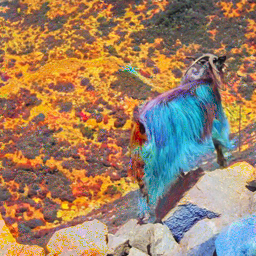}
        \caption{w/o changing prompt - inappropriate colors}
        \label{fig:fixed_saturated}
    \end{subfigure}
\vspace{-1em} 
\caption{Example of frame colorization with and without changing prompt every $\Delta t$ and change of scene in DAVIS30 pairs.}
\label{fig:prompt_comparison}
\vspace{-4em}
\end{figure}

Using dynamic detailed prompts significantly improves PSNR (29.42 vs 28.23) and slightly improves temporal consistency (CDC ratio 1.15 vs 1.22). While the fixed-prompt version shows a slightly better Colorfulness ratio, qualitative results, Figure \ref{fig:prompt_comparison} highlight the importance of adapting guidance. A fixed detailed prompt can lead desaturated or incorrect colors as the scene content diverges from the initial description (Fig \ref{fig:fixed_gray}), or conversely, overly saturated and unnatural colors if the initial prompt locks onto specific colors that are no longer appropriate (Fig \ref{fig:fixed_saturated}). Dynamically updated detailed prompts allow the colorization to adapt to evolving content, leading to more accurate and plausible results overall when such guidance is available.
    
\subsubsection{Impact of Warping Correction}
\begin{figure}[!htb]
    \centering

    \begin{subfigure}[b]{0.4\textwidth}
        \centering
        \includegraphics[width=.7\linewidth]{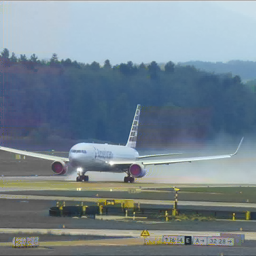}
        \caption{With warping correction}
        \label{fig:with_correction_plane}
    \end{subfigure}
    \hspace{20px} 
    \begin{subfigure}[b]{0.4\textwidth}
        \centering
        \includegraphics[width=.7\linewidth]{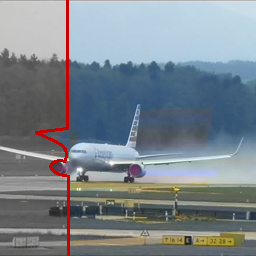}
        \caption{Without warping correction}
        \label{fig:no_correction_plane}
    \end{subfigure}

    \caption{Example of frame colorization with and without warping correction.}
    \label{fig:warping_comparison}
\end{figure}

We evaluate the necessity of our warping correction mechanism by comparing the full method against a version where the warped frame $I_{warp}(t+1)$ is used directly as the final output, without correcting corrupted areas identified by Eq. \ref{eq:corr_mask}. Table \ref{tab:warping correction ablation} shows the results. Without correction, PSNR and CDC scores are slightly better on average. However, the Colorfulness ratio drops dramatically (0.65 vs 0.94 on DAVIS, 0.66 vs 1.05 on Videvo), indicating significantly desaturated results. Qualitative examples in Figure \ref{fig:warping_comparison} clearly illustrate why correction is crucial. Without it, areas with occlusions, disocclusions, or complex motion where optical flow fails remain uncolored (appearing gray) or exhibit severe color shifting artifacts (Fig \ref{fig:no_correction_plane}). The correction mechanism (Eq. \ref{eq:final_comp}) effectively fills these regions using the current frame's L-CAD prediction, restoring plausible colors and significantly improving the visual quality and vibrancy (Fig \ref{fig:with_correction_plane}), even if the filled colors slightly deviate from the ground truth, minimally impacting average PSNR/CDC. This demonstrates that the correction step is essential for perceptual quality and avoiding distracting artifacts.
\begin{table}[!htb]
\vspace{-10pt}
\centering
\resizebox{.9\textwidth}{!}{%
    \begin{tabular}{lccc|ccc}
    \hline
     & \multicolumn{3}{c}{\textbf{DAVIS30}} & \multicolumn{3}{c}{\textbf{VIDEVO20}} \\ \hline
    {\textbf{Method}} 
    & $\frac{\text{CDC}}{\text{True CDC}}^*$  & PSNR$\uparrow$ &  $\frac{\text{Colorfulness}}{\text{True Colorfulness}}^*$    & $\frac{\text{CDC}}{\text{True CDC}}^*$  &PSNR$\uparrow$ &  $\frac{\text{Colorfulness}}{\text{True Colorfulness}}^*$  \\ \hline
    Without warping correction & \textbf{1.25} & \textbf{29.64} & 0.65 & \textbf{0.98} & \textbf{30.62} & 0.66 \\ 
    With warping correction  & 1.25 & 29.63 & \textbf{0.94} & 1.08 & 30.26 & \textbf{1.05} \\ \hline
    \end{tabular}
}
\caption{Comparison between the proposed method and its version without correcting gray areas. $^*$ Best value is closer to 1.}
\label{tab:warping correction ablation}
\end{table}
\vspace{-20pt}
\section{Conclusions}\label{sec:conclusions}
This paper introduced a novel framework for automatic video colorization that leverages language-based guidance and robust temporal consistency enforcement. By combining a language-conditioned diffusion model with automatically generated object masks and textual prompts, our method achieves detailed and plausible colorization. Temporal coherence is maintained through optical flow-based color propagation combined with a crucial correction step that addresses warping artifacts, ensuring smooth and visually pleasing results. 
We report state-of-the-art performance on standard benchmarks using only a generic prompt, significantly improving reconstruction fidelity and color vibrancy over previous automatic methods while maintaining competitive temporal consistency. Furthermore, we show that incorporating detailed, color-specific prompts (simulating user guidance) further enhances performance and allows for semantic control over the colorization process.

\paragraph{Acknowledgments}
Work partially supported by the Italian Ministry of Education and Research (MUR) in the framework of the FoReLab project (Departments of Excellence).

\vspace{-10pt}
\bibliographystyle{plain}
\bibliography{iciap25}

\end{document}